\PassOptionsToPackage{colorlinks=true,linkcolor=blue,citecolor=blue,urlcolor=blue,breaklinks=true}{hyperref}
\documentclass[runningheads]{llncs}
\usepackage{tikz}
\usetikzlibrary{positioning, arrows.meta, shapes.geometric}
\usepackage[T1]{fontenc}
\usepackage{lmodern}
\usepackage{microtype}
\usepackage{graphicx}
\usepackage{booktabs}
\usepackage{tabularx}
\usepackage{array}
\usepackage{adjustbox}
\usepackage{multirow}
\usepackage{url}
\usepackage{amsmath}
\usepackage{amssymb}
\usepackage{placeins}
\usepackage{float}
\usepackage{orcidlink}
\usepackage{xcolor}
\usepackage{capt-of}
\usepackage{tikz}
\usetikzlibrary{shapes,arrows.meta,positioning,fit,backgrounds,calc,shadows.blur}
\usepackage{tikz}
\usetikzlibrary{positioning, arrows.meta}
\usepackage{xcolor}
\definecolor{tierData}{HTML}{D6E4F0}
\definecolor{tierRep}{HTML}{D9EAD3}
\definecolor{tierEval}{HTML}{FCE5CD}
\definecolor{tierMeta}{HTML}{EFEFEF}
\definecolor{borderData}{HTML}{4A6FA5}
\definecolor{borderRep}{HTML}{4F8A4A}
\definecolor{borderEval}{HTML}{C97A3F}
\definecolor{borderMeta}{HTML}{888888}

\usepackage[colorlinks=true,linkcolor=blue,citecolor=blue,urlcolor=blue,breaklinks=true]{hyperref}

\begin{document}

\title{TinyCNN: A 193K-Parameter Network for On-Device Plant Disease Detection, with a Cross-Dataset Robustness Diagnosis}
\titlerunning{TinyCNN}

 \author{ 
 Ngoc-Bao Ho-Lam\inst{1,2}\orcidlink{0009-0003-2315-9903} \and
 Thai-Anh Nguyen\inst{3}\orcidlink{0009-0005-5600-6510}}
% %
\titlerunning{TinyCNN for On-Device Plant Disease Detection}
 \authorrunning{Ho-Lam and Nguyen}
 \institute{
 University of Science, VNU-HCM, Vietnam \and
 Vietnam National University, Ho Chi Minh City, Vietnam \and
 Faculty of Information Technology, Van Lang School of Technology, Van Lang University, Ho Chi Minh City, Vietnam \\
 \email{hlnbao23@apcs.fitus.edu.vn} \\
 \email{anh.nt@vlu.edu.vn}}

\maketitle

\begin{abstract}
Detecting crop disease early is central to sustainable agriculture and food security under United Nations Sustainable Development Goal~2 (Zero Hunger), and is especially urgent in resource-constrained regions where expert diagnosis is scarce but low-cost mobile devices are widespread. This paper presents TinyCNN, a lightweight convolutional neural network for on-device plant disease classification. TinyCNN uses depthwise separable convolution blocks and contains only 193,190 trainable parameters with 110.05M MACs for a $224 \times 224$ input image. On the 38-class PlantVillage benchmark, TinyCNN achieves 98.88\% test accuracy and 98.03\% macro-F1 while being approximately 58$\times$ smaller than ResNet18 and 11.8$\times$ smaller than a MobileNetV2 teacher, directly reducing the energy, memory, and cost footprint of inference in line with Green AI principles. The paper further analyzes vanilla knowledge distillation as a sustainable model-compression strategy; an ablation over $\alpha \in \{0.3,0.5,0.7\}$ and $T \in \{2,4\}$ selects $\alpha=0.3$, $T=4$, producing a distilled TinyCNN with 98.81\% test accuracy. Finally, cross-dataset evaluation from PlantVillage to PlantDoc reveals a substantial robustness gap under real-world conditions, which a Grad-CAM analysis attributes to off-leaf, background-driven attention consistent with shortcut learning. TinyCNN is thus an energy-efficient, deployable building block for sustainable agricultural intelligence, while field robustness remains the key barrier to durable real-world impact.

\end{abstract}

\section{Introduction}

Plant diseases are a major threat to agricultural productivity and food security, causing substantial yield losses worldwide \cite{savary2019global}. Containing them is therefore central to the United Nations Sustainable Development agenda, contributing directly to Zero Hunger (SDG~2) and, through resource-efficient computation, to Responsible Consumption and Production (SDG~12), with reduced operational energy serving as a proxy for the carbon reductions targeted by Climate Action (SDG~13). Early, accurate diagnosis matters most in rural and resource-constrained regions, where expert plant pathologists are scarce but low-cost smartphones are increasingly ubiquitous. Such settings call for first-line visual diagnosis that runs \emph{on the device itself}, which in turn demands models that are accurate, compact, and energy-efficient rather than dependent on cloud connectivity or high-end hardware.

Recent deep models reach strong accuracy on benchmarks such as PlantVillage \cite{hughes2015open}, but accuracy alone is insufficient for \emph{sustainable} deployment: large CNNs demand memory, computation, and energy that exceed low-resource devices and inflate the carbon footprint of inference \cite{schwartz2020green,strubell2019energy}. Plant disease detection should therefore be treated jointly as a classification and a sustainability/deployment problem.

This paper proposes TinyCNN, a compact network that factorizes standard convolution into depthwise and pointwise operations \cite{howard2017mobilenets,chollet2017xception} to reach 98.88\% accuracy and 98.03\% macro-F1 on the 38-class PlantVillage benchmark with only 193,190 parameters and 110.05M MACs, making it approximately 58$\times$ smaller than ResNet18 and 11.8$\times$ smaller than the MobileNetV2 teacher at lower CPU latency.

We also analyze knowledge distillation (KD), which trains a small student from the hard labels and the softened predictions of a larger teacher \cite{hinton2015distilling}, as a complementary compression strategy. KD need not improve a student when the teacher is near-saturated or the teacher-student capacity gap is large \cite{mirzadeh2020improved}; we call a dataset on which both models already reach near-ceiling accuracy a \emph{saturated benchmark}. Ablating the distillation weight $\alpha$ and temperature $T$ and selecting $\alpha=0.3,T=4$, our distilled TinyCNN matches but does not exceed the scratch-trained model, indicating limited benefit once the student already has sufficient capacity.

Finally, cross-dataset evaluation on PlantDoc \cite{singh2020plantdoc} shows that all models, though strong on PlantVillage, degrade sharply under real-world conditions (e.g., MobileNetV2 falls from 99.77\% to 28.57\%), confirming that high in-distribution accuracy does not guarantee field reliability.

The main contributions of this paper are as follows:
\begin{itemize}
    \item \textbf{Lightweight architecture.} We propose TinyCNN, a 193K-parameter network for on-device plant disease classification that reaches 98.88\% accuracy on PlantVillage while being $\sim$58$\times$ smaller than ResNet18.
    \item \textbf{A quantitative shortcut-learning diagnosis.} We introduce an \emph{off-leaf ratio} and use Grad-CAM to show that PlantVillage-trained models attend predominantly to background rather than disease tissue, \emph{even when their predictions are correct}, giving a mechanistic explanation for their cross-dataset collapse.
    \item \textbf{A knowledge distillation insight.} We show that on a near-saturated benchmark vanilla logit-based KD acts only as a tunable regularizer, and that $\alpha$ and $T$ interact and must be tuned jointly rather than independently.
\end{itemize}
\section{Related Work}

\subsection{Deep Learning for Plant Disease Detection}

Deep learning has become a dominant approach for image-based plant disease recognition. PlantVillage is one of the most widely used datasets in this area, providing labeled images of healthy and diseased plant leaves across multiple crop species \cite{hughes2015open}. Many CNN-based methods have reported very high accuracy on this benchmark, demonstrating the effectiveness of convolutional feature extraction for disease classification \cite{mohanty2016using,too2019comparative,ferentinos2018deep}. However, PlantVillage images are mostly captured under controlled conditions, often with clean backgrounds and consistent image composition. As a result, high PlantVillage accuracy may overestimate real-world performance.

To address this limitation, PlantDoc was introduced as a more challenging benchmark containing plant disease images captured in natural environments \cite{singh2020plantdoc}. These images include complex backgrounds, varying illumination, occlusion, and uncontrolled camera conditions. Cross-dataset evaluation from PlantVillage to PlantDoc is therefore important for measuring whether a model learns robust disease-related features or relies on dataset-specific visual patterns. Prior work has documented this PlantVillage-to-PlantDoc accuracy drop; the present paper instead \emph{explains its mechanism}, using Grad-CAM to show quantitatively that the collapse coincides with model attention drifting off the leaf onto background context.

\subsection{Lightweight CNNs and Edge AI}

Efficient neural network design is essential for edge AI and sustainable computing. MobileNets introduced depthwise separable convolution to reduce computation while maintaining competitive accuracy \cite{howard2017mobilenets}. MobileNetV2 further improved this design using inverted residuals and linear bottlenecks \cite{sandler2018mobilenetv2}. Other compact architectures, such as ShuffleNet and EfficientNet, also explore accuracy-efficiency trade-offs through channel shuffling, compound scaling, or optimized architecture search \cite{zhang2018shufflenet,tan2019efficientnet}.

In agriculture, deployment on smartphones, embedded boards, or low-power monitors makes such efficiency essential, and Green AI further motivates smaller, lower-energy models \cite{schwartz2020green,strubell2019energy}. TinyCNN follows this direction with a task-specific compact design rather than a larger general-purpose CNN.

\subsection{Knowledge Distillation}

Knowledge distillation was popularized by Hinton \textit{et al.}, who showed that a student model can learn from the softened output distribution of a teacher model \cite{hinton2015distilling}. The teacher distribution may contain ``dark knowledge'', such as similarity between classes, which is not captured by one-hot labels. KD has since been applied to many model compression and transfer learning tasks \cite{gou2021knowledge,romero2015fitnets,passalis2018learning}.

However, KD performance depends strongly on the teacher, student, data distribution, and hyperparameters. Mirzadeh \textit{et al.} showed that a large teacher-student capacity gap can make distillation less effective, motivating teacher-assistant distillation \cite{mirzadeh2020improved}. Other studies have also shown that stronger teachers are not always better teachers for small students \cite{cho2019efficacy}. In the context of PlantVillage, where both teacher and student can reach near-ceiling accuracy, KD must be analyzed carefully rather than assumed to improve performance.

\section{Method}

\subsection{TinyCNN Architecture}

TinyCNN is designed as a compact convolutional neural network for 38-class plant disease classification under edge-device constraints. The model takes an RGB image of size $224 \times 224$ as input and produces a probability distribution over plant disease classes. The main design goal is to obtain a favorable trade-off between accuracy and computational efficiency by reducing the number of parameters and multiply-accumulate operations (MACs), while still preserving enough representational capacity for fine-grained disease symptoms.

\begin{figure}[h!]
\centering
\includegraphics[width=\columnwidth]{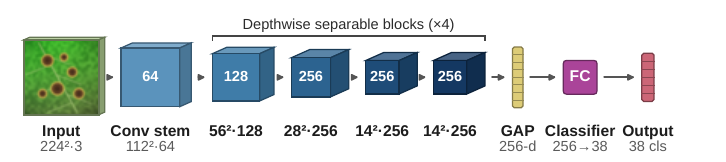}
\caption{Architecture of the proposed TinyCNN for on-device plant disease detection. A $3\times3$ convolutional stem is followed by four depthwise separable convolution blocks, adaptive global average pooling, and a lightweight linear classifier. Block height encodes spatial resolution and block depth encodes channel count. The model has 193,190 trainable parameters and 110.05M MACs for a $3\times224\times224$ input.}
\label{fig:tinycnn_architecture}
\end{figure}

The overall architecture follows a simple hierarchical CNN design. It starts with a standard convolutional stem that extracts low-level visual patterns such as edges, textures, and color variations. The stem is followed by four depthwise separable convolution blocks. These blocks progressively reduce the spatial resolution while increasing the channel dimension, allowing the network to learn more abstract disease-related features at lower computational cost. Finally, global average pooling is used to aggregate spatial information, followed by dropout and a fully connected classifier. Fig.~\ref{fig:tinycnn_architecture} illustrates the overall TinyCNN architecture.

The model can be summarized as:
\begin{equation}
x \rightarrow \text{Conv Stem} \rightarrow \text{DSConv Blocks} \rightarrow \text{GAP} \rightarrow \text{Dropout} \rightarrow \text{FC}.
\end{equation}

The convolutional stem consists of a $3 \times 3$ convolution from 3 input channels to 64 channels with stride 2, followed by batch normalization and ReLU6 activation. ReLU6 is used because it is commonly adopted in mobile-oriented architectures \cite{howard2017mobilenets,sandler2018mobilenetv2} and provides bounded activation values, which can be beneficial for efficient deployment and later quantization.

Each depthwise separable block factorizes a standard convolution into a depthwise $3\times3$ convolution (one spatial filter per channel) and a pointwise $1\times1$ convolution (cross-channel mixing), each followed by batch normalization and ReLU6; this factorization sharply reduces computation. The four blocks are configured as $64\!\rightarrow\!128$ (stride 2), $128\!\rightarrow\!256$ (stride 2), $256\!\rightarrow\!256$ (stride 2), and $256\!\rightarrow\!256$ (stride 1): the first three reduce spatial resolution while raising the semantic level, and the last refines the high-level features. Adaptive global average pooling then produces a 256-dimensional vector, and a dropout layer ($p{=}0.2$) precedes the final linear classifier. The complete model contains 193,190 trainable parameters and 110.05M MACs for a $224\times224$ input, making it suitable for on-device deployment.

\subsection{Teacher and Baseline Models}

To evaluate TinyCNN in a deployment-oriented setting, we compare it with two pretrained convolutional architectures: MobileNetV2 and ResNet18. Both models are initialized with ImageNet-pretrained weights, following the common transfer learning practice in plant disease recognition, where pretrained visual features can accelerate convergence and improve performance when fine-tuned on agricultural image datasets \cite{too2019comparative,ferentinos2018deep}.

MobileNetV2, our teacher, balances accuracy and efficiency while remaining substantially larger than TinyCNN; it is built on inverted residual blocks and linear bottlenecks \cite{sandler2018mobilenetv2}. We replace its ImageNet classifier with a task-specific head:
\begin{equation}
\text{Dropout}(0.2) \rightarrow \text{Linear}(1280,38).
\end{equation}
The resulting teacher has 2,272,550 parameters and 326.26M MACs, making it a strong yet efficient source of soft-label supervision for the TinyCNN student.

ResNet18, a standard high-capacity residual baseline \cite{he2016deep}, is not used as a teacher but as a reference for the efficiency--accuracy frontier. Its ImageNet head is likewise replaced with a 38-class layer, giving 11,196,006 parameters and 1.82B MACs.

Together, these baselines let us assess whether a compact student can approach a stronger mobile-oriented teacher (MobileNetV2) and how TinyCNN sits on the efficiency--accuracy frontier relative to a conventional high-capacity CNN (ResNet18).

\subsection{Knowledge Distillation Objective}

Knowledge distillation transfers predictive information from the frozen, PlantVillage-fine-tuned MobileNetV2 teacher to the TinyCNN student. For each image, the teacher and student produce logits $z_t$ and $z_s$, and the student is trained with two signals: a cross-entropy loss on the hard label $y$, and a distillation loss that matches the teacher's softened class distribution. For a temperature value $T$, the softened probability of class $i$ is defined as:
\begin{equation}
p_i^T(z)=
\frac{\exp(z_i/T)}
{\sum_j \exp(z_j/T)}.
\end{equation}

A larger temperature produces a softer probability distribution, making non-maximum classes more visible. These secondary probabilities can encode inter-class similarity, such as visually similar plant diseases, and are commonly referred to as dark knowledge \cite{hinton2015distilling}. The complete distillation objective is:
\begin{equation}
\mathcal{L}_{KD}
=
(1-\alpha)\mathcal{L}_{CE}(z_s,y)
+
\alpha T^2
\mathcal{L}_{KL}
\left(
p^T(z_t) \parallel p^T(z_s)
\right).
\end{equation}

Here, $\alpha$ weights the teacher supervision and $T$ controls its smoothness; the factor $T^2$ preserves the gradient scale across temperatures \cite{hinton2015distilling}. A larger $\alpha$ makes the student rely more on the teacher, while a smaller $\alpha$ keeps the ground-truth labels as the dominant signal.

\section{Experimental Setup}

\subsection{Datasets}

The in-distribution dataset is PlantVillage \cite{hughes2015open}, a widely used benchmark of leaf images across multiple crop species and disease categories. We merge the available training and validation folders into a single set of 54,305 images across 38 classes, then re-split it with a stratified 70:15:15 ratio into 38,013 training, 8,146 validation, and 8,146 test images, preserving the class distribution across splits.

\begin{figure}[t]
\centering
\includegraphics[width=\textwidth]{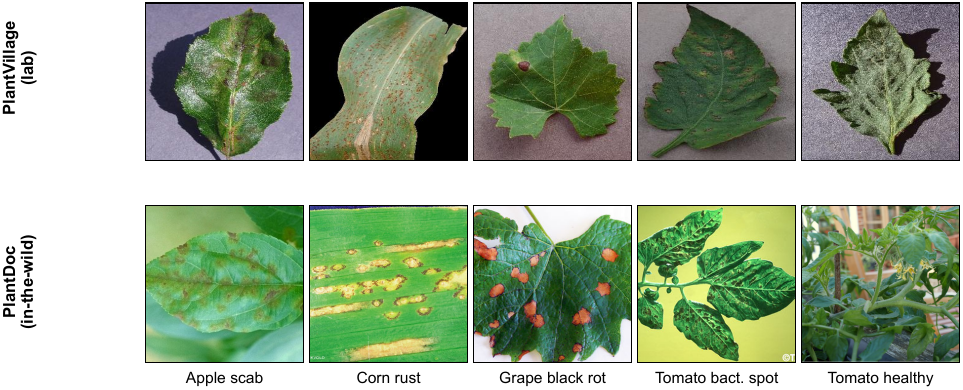}
\caption{Visual comparison between PlantVillage (top, controlled laboratory conditions) and PlantDoc (bottom, in-the-wild field conditions) for five representative classes. PlantDoc images exhibit complex backgrounds, variable illumination, occlusion, and uncontrolled composition, which motivates the cross-dataset robustness evaluation in Section~\ref{sec:plantdoc}.}
\label{fig:domain_shift}
\end{figure}

For out-of-distribution evaluation, we use PlantDoc \cite{singh2020plantdoc}, which contains plant disease images captured under more realistic field conditions. Unlike PlantVillage, PlantDoc includes complex backgrounds, variable lighting, and uncontrolled image composition, making it suitable for evaluating cross-dataset robustness. In this study, we use the PlantDoc test set containing 252 images across 27 mapped classes. Each PlantDoc class is manually mapped to its corresponding PlantVillage class. Fig.~\ref{fig:domain_shift} illustrates the visual gap between the two datasets for several representative classes.

Two evaluation settings are used for PlantDoc. In the strict setting, the model predicts over all 38 PlantVillage classes, which tests whether the model can select the correct mapped class without any restriction. In the restricted setting, predictions are limited to the mapped PlantVillage classes only, which evaluates performance after removing unmapped output classes from consideration.

\subsection{Data Preprocessing and Augmentation}

All images are resized to $224 \times 224$ for compatibility across TinyCNN, MobileNetV2, and ResNet18, matching the input of the ImageNet-pretrained backbones \cite{sandler2018mobilenetv2,he2016deep}. For training, each image is resized so its shorter side is $\lfloor 1.15 \times 224 \rfloor$ and then randomly resized-cropped with scale $(0.7, 1.0)$, followed by random horizontal flipping ($p{=}0.5$), random rotation within $\pm 15^\circ$, and color jitter (brightness, contrast, saturation $=0.2$) to simulate field variation in orientation, lighting, and capture conditions \cite{shorten2019survey}. For validation and testing, images are resized and center-cropped to $224 \times 224$ for deterministic evaluation. All images are normalized with the ImageNet mean and standard deviation, consistent with the pretrained backbones \cite{deng2009imagenet}.

\subsection{Training Configuration}

All models follow a consistent protocol for fair comparison. We use AdamW (weight decay $10^{-4}$) \cite{loshchilov2019decoupled}, a cosine-annealing schedule \cite{loshchilov2017sgdr}, label smoothing $0.1$ \cite{szegedy2016rethinking}, and gradient clipping at maximum norm $1.0$ \cite{pascanu2013difficulty}, with batch size 96 and a fixed seed of 42. The ImageNet-pretrained MobileNetV2 and ResNet18 are fine-tuned at learning rate $10^{-4}$, whereas TinyCNN is trained from scratch at $10^{-3}$.

The MobileNetV2 teacher, ResNet18 baseline, scratch-trained TinyCNN, and final distilled TinyCNN are each trained for 15 epochs; each candidate in the KD ablation is trained for 8 epochs to compare $(\alpha,T)$ settings efficiently before final retraining.

\subsection{Evaluation Metrics}

We report accuracy and macro-F1 on the PlantVillage test set; macro-F1 weights all classes equally and is important under the class imbalance typical of disease datasets \cite{sokolova2009systematic,powers2011evaluation}. For deployment efficiency we report trainable parameters, multiply-accumulate operations (MACs), state-dict size, and CPU latency, the last measured at batch size 1 to approximate single-image on-device inference \cite{howard2017mobilenets,sandler2018mobilenetv2}. For cross-dataset robustness we additionally report strict and restricted accuracy and macro-F1 on PlantDoc.

\section{Results}

\subsection{PlantVillage Performance}

Table~\ref{tab:plantvillage} reports the PlantVillage test performance and efficiency metrics. MobileNetV2 achieves the highest accuracy, with 99.77\% accuracy and 99.65\% macro-F1. ResNet18 obtains 99.67\% accuracy and 99.49\% macro-F1. TinyCNN trained from scratch achieves 98.88\% accuracy and 98.03\% macro-F1 with only 193K parameters.

\begin{table*}[h!]
\centering
\caption{PlantVillage test performance and efficiency comparison.}
\label{tab:plantvillage}
\small
\begin{tabular}{lcccccc}
\hline
\textbf{Model} & \textbf{Params} & \textbf{MACs} & \textbf{Size} & \textbf{Mean Lat.} & \textbf{Acc.} & \textbf{Macro-F1} \\
\hline
MobileNetV2 teacher & 2.27M & 326.26M & 26.36 MB & 17.95 ms & 99.77 & 99.65 \\
ResNet18 & 11.20M & 1823.54M & 128.25 MB & 16.75 ms & 99.67 & 99.49 \\
TinyCNN scratch & 0.19M & 110.05M & 2.26 MB & 4.27 ms & 98.88 & 98.03 \\
TinyCNN KD $(\alpha=0.3,T=4)$ & 0.19M & 110.05M & 2.26 MB & 3.92 ms & 98.81 & 97.72 \\
\hline
\end{tabular}
\end{table*}

The results show that TinyCNN provides a strong efficiency-accuracy trade-off. Compared with ResNet18, TinyCNN is approximately 58$\times$ smaller in parameters and requires about 16.6$\times$ fewer MACs; compared with MobileNetV2, it is 11.8$\times$ smaller and 4.2$\times$ faster on CPU, at a cost of only 0.89 percentage points of accuracy. MobileNetV2 is slightly slower than ResNet18 on CPU despite far fewer MACs, which is expected because its depthwise convolutions are memory-bound and less cache-friendly. Since the scratch-trained and distilled TinyCNN share an identical architecture, their inference cost is effectively the same; the small latency difference is measurement variance, not a speed-up from distillation.

\subsection{Knowledge Distillation Hyperparameter Analysis}

To determine the final distillation setting, we conduct a small ablation study over the distillation weight and temperature. Specifically, we evaluate 
$\alpha \in \{0.3,0.5,0.7\}$ and $T \in \{2,4\}$. Each candidate configuration is trained for 8 epochs and compared using validation accuracy and macro-F1. The best validation configuration is then selected and retrained for 15 epochs to obtain the final distilled TinyCNN model.

\begin{table}[h!]
\centering
\caption{Knowledge distillation hyperparameter ablation. Each configuration is trained for 8 epochs and selected based on validation performance.}
\label{tab:kd_ablation}
\begin{tabular}{ccccc}
\hline
$\alpha$ & $T$ & Val. Acc. & Val. F1 & Test Acc. \\
\hline
0.3 & 2 & 94.00 & 91.32 & 94.00 \\
0.3 & 4 & \textbf{97.62} & \textbf{96.16} & \textbf{97.86} \\
0.5 & 2 & 97.46 & 95.64 & 97.46 \\
0.5 & 4 & 97.43 & 95.04 & 97.54 \\
0.7 & 2 & 97.13 & 94.55 & 97.14 \\
0.7 & 4 & 96.98 & 93.85 & 97.29 \\
\hline
\end{tabular}
\end{table}

As shown in Table~\ref{tab:kd_ablation}, the best validation result is obtained with $\alpha=0.3$ and $T=4$, reaching 97.62\% validation accuracy and 96.16\% validation macro-F1 during the ablation stage. We caution against reading this as a simple ``lower $\alpha$ is always better'' trend, because the effect of $\alpha$ is not monotone and interacts with the temperature $T$. At $T=4$, decreasing $\alpha$ from 0.7 to 0.3 improves validation accuracy ($96.98\% \rightarrow 97.62\%$), consistent with keeping the ground-truth cross-entropy as the dominant learning signal. At $T=2$, however, the same low-$\alpha$ setting is the weakest configuration ($94.00\%$), while the intermediate $\alpha=0.5$ performs best. The most plausible explanation is that a low distillation weight becomes beneficial only once the teacher distribution has been sufficiently softened: at $T=2$ the soft targets remain sharp and close to one-hot, so a small $\alpha$ provides too weak a learning signal within the short 8-epoch ablation budget. This also accounts for the comparatively low $94.00\%$ outlier at $(\alpha=0.3,T=2)$, and indicates that $\alpha$ and $T$ should be tuned jointly rather than independently.

This behavior is consistent with previous observations that knowledge distillation is sensitive to the teacher-student capacity relationship and does not always improve a compact student model \cite{mirzadeh2020improved,cho2019efficacy,gou2021knowledge}. In our case, the MobileNetV2 teacher reaches 99.77\% test accuracy on PlantVillage while the scratch-trained TinyCNN already reaches 98.88\%, so the student appears to have enough capacity to learn the main discriminative patterns from hard labels alone.

After selecting $\alpha=0.3,T=4$, we retrain the final distilled TinyCNN for 15 epochs. It reaches 98.81\% test accuracy and 97.72\% macro-F1, essentially matching the scratch-trained TinyCNN (98.88\%, 98.03\%). Under a single fixed seed, a 0.07-point gap lies within run-to-run variance, so the conservative conclusion is that vanilla KD yields no measurable improvement on this saturated benchmark: even at $T=4$, which softens the teacher and exposes its dark knowledge, the soft targets add little beyond the one-hot labels because teacher and student already agree on almost all examples. For compact models near the dataset ceiling, architecture design thus matters more than teacher supervision, and further gains likely require feature-level distillation, attention transfer, or more challenging training data.

\subsection{Cross-Dataset Evaluation on PlantDoc}
\label{sec:plantdoc}

Table~\ref{tab:plantdoc} reports cross-dataset evaluation from PlantVillage to PlantDoc. The results reveal a large domain shift. MobileNetV2 achieves the best strict accuracy, but only reaches 28.57\%. ResNet18 achieves 26.98\%, while TinyCNN scratch obtains 13.10\%. Among KD ablation checkpoints, $\alpha=0.7,T=2$ achieves the best strict and restricted PlantDoc accuracy, but the absolute performance remains low. Notably, the configuration that maximizes in-distribution validation accuracy ($\alpha=0.3,T=4$) is not the one with the highest out-of-distribution accuracy ($\alpha=0.7,T=2$); these PlantDoc differences are small, but the pattern cautions that in-distribution model selection need not track field robustness. For transparency, the KD rows are 8-epoch ablation checkpoints whereas the baselines are fully trained (15-epoch) networks.

\begin{table*}[h!]
\centering
\caption{Cross-dataset evaluation on PlantDoc.}
\label{tab:plantdoc}
\begin{tabular}{lcccc}
\hline
Model & Strict Acc. & Strict Macro-F1 & Restricted Acc. & Restricted Macro-F1 \\
\hline
MobileNetV2 teacher & 28.57 & 20.73 & 31.35 & 27.64 \\
ResNet18 & 26.98 & 20.94 & 30.16 & 28.09 \\
TinyCNN scratch & 13.10 & 8.94 & 17.06 & 14.29 \\
TinyCNN KD $(\alpha=0.3,T=2)$ & 10.71 & 7.10 & 14.68 & 11.66 \\
TinyCNN KD $(\alpha=0.3,T=4)$ & 13.10 & 9.91 & 15.08 & 12.57 \\
TinyCNN KD $(\alpha=0.5,T=2)$ & 10.32 & 7.83 & 16.27 & 14.86 \\
TinyCNN KD $(\alpha=0.5,T=4)$ & 8.33 & 5.92 & 12.70 & 10.61 \\
TinyCNN KD $(\alpha=0.7,T=2)$ & 13.89 & 11.46 & 17.86 & 15.70 \\
TinyCNN KD $(\alpha=0.7,T=4)$ & 13.49 & 9.50 & 17.06 & 14.12 \\
\hline
\end{tabular}
\end{table*}

The PlantDoc results show that strong in-distribution performance does not guarantee field robustness. Across the $n{=}252$ mapped images, MobileNetV2 falls from 99.77\% PlantVillage accuracy to 28.57\% strict PlantDoc accuracy (95\% Wilson interval $23.3$--$34.4\%$) and TinyCNN from 98.88\% to 13.10\% ($9.5$--$17.8\%$); the resulting gaps of $71.20$ and $85.78$ percentage points far exceed these intervals. Because the test set is small, within-group differences warrant caution: the two strongest models overlap (ResNet18 $21.9$--$32.8\%$ vs.\ MobileNetV2 $23.3$--$34.4\%$) and the KD-checkpoint rows in Table~\ref{tab:plantdoc} differ within overlapping Wilson intervals, so the robust finding is the uniformly low absolute accuracy rather than the rank order of individual rows.

\begin{figure}[t]
\centering
\includegraphics[width=\textwidth]{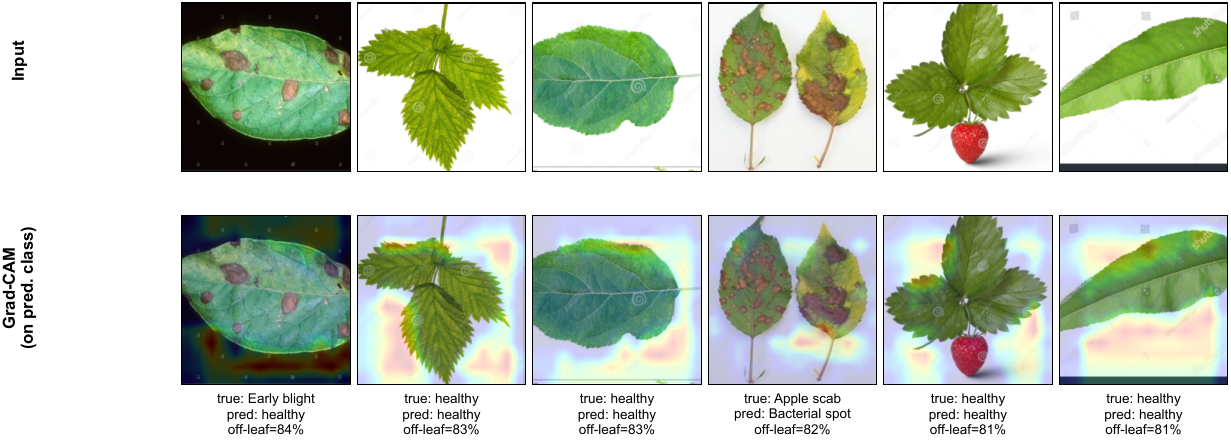}
\caption{Grad-CAM error analysis on PlantDoc. The top row shows input images and the bottom row shows Grad-CAM saliency for the predicted class. In every case the off-leaf ratio (the fraction of activation mass outside the leaf region) exceeds 80\%, revealing that the model attends to background context rather than disease symptoms. High off-leaf attention appears both in misclassified samples and in correctly classified ``healthy'' samples, indicating reliance on spurious, dataset-specific cues (shortcut learning) rather than transferable disease features.}
\label{fig:gradcam_failures}
\end{figure}

To understand \emph{why} performance collapses under domain shift, we perform a Grad-CAM error analysis \cite{selvaraju2017gradcam} of the predicted class on representative PlantDoc images. As Fig.~\ref{fig:gradcam_failures} shows, attention frequently drifts off the leaf onto the background; we summarize this with an \emph{off-leaf ratio}, the fraction of Grad-CAM activation mass outside the leaf region, which exceeds 80\% across the displayed cases. Crucially, high off-leaf attention appears not only in misclassifications (e.g., Early blight read as healthy) but also in correctly classified ``healthy'' samples, so the model can be right for the wrong reasons. This is characteristic of shortcut learning \cite{geirhos2020shortcut}, where a model exploits dataset-specific cues such as the uniform laboratory backgrounds of PlantVillage that do not transfer to cluttered field conditions, giving a mechanistic explanation for the gap in Table~\ref{tab:plantdoc}: the models have partly learned to recognize acquisition conditions rather than disease symptoms.

\subsection{Discussion}

Taken together, the experiments yield three deployment-oriented messages. (i)~A task-specific compact architecture suffices on a controlled benchmark: TinyCNN matches much larger backbones on PlantVillage while using $11.8\times$ fewer parameters and $\sim3\times$ less per-image CPU compute than the MobileNetV2 teacher, directly lowering the energy and memory footprint of inference and making continuous monitoring feasible on battery- or solar-powered edge hardware \cite{howard2017mobilenets,sandler2018mobilenetv2,schwartz2020green}; as a first-order estimate, at the measured $4.27$\,ms CPU latency and a representative edge-CPU draw of a few watts, a single inference costs on the order of ten millijoules (an order-of-magnitude figure, not a calibrated measurement), so the $4.2\times$ latency reduction over MobileNetV2 translates into a comparable per-inference energy saving. (ii)~On a near-saturated benchmark, vanilla logit-based KD behaves only as a tunable regularizer: with a sufficiently capable student, the soft labels add little beyond the hard labels \cite{mirzadeh2020improved,cho2019efficacy,gou2021knowledge}. (iii)~In-distribution accuracy is not a proxy for field reliability: all models collapse on PlantDoc, and Grad-CAM traces this to off-leaf, background-driven attention, a form of shortcut learning that does not transfer across domains \cite{selvaraju2017gradcam,geirhos2020shortcut}. Closing this gap will require domain generalization and field-oriented training rather than larger backbones \cite{shorten2019survey,zhou2022domain}.

\textbf{Limitations.} Three caveats temper these conclusions. All results use a single training seed, so we report differences only when they exceed plausible run-to-run variance; multi-seed estimates would strengthen the KD comparison. The mapped PlantDoc set is small (252 images, 27 classes), so its numbers measure domain-shift sensitivity rather than full field performance. Finally, the off-leaf ratio depends on the leaf segmentation used to define the leaf region and is reported as a qualitative diagnostic.

\section{Conclusion}

This paper presented TinyCNN, a 193,190-parameter depthwise-separable network for sustainable on-device plant disease detection that reaches 98.88\% accuracy and 98.03\% macro-F1 on the 38-class PlantVillage benchmark while being roughly 58$\times$ smaller than ResNet18, making it well suited to low-resource edge deployment.

This paper also analyzed vanilla knowledge distillation with a MobileNetV2 teacher. An ablation over $\alpha$ and $T$ selected $\alpha=0.3,T=4$, but the distilled model is statistically indistinguishable from the scratch-trained TinyCNN: on a near-saturated benchmark with a sufficiently capable student, vanilla logit-based distillation provides no measurable benefit, and $\alpha$ and $T$ must be tuned jointly.

Finally, cross-dataset evaluation on PlantDoc revealed a substantial robustness gap that Grad-CAM traces to off-leaf, background-driven attention: strong performance on controlled datasets does not imply field reliability. Future work should prioritize domain generalization, field-oriented augmentation, on-device quantization (ONNX/TFLite), and multi-seed validation on real edge hardware.

\section*{Acknowledgments}

This research is supported by research funding from Faculty of Information Technology, University of Science, Vietnam National University - Ho Chi Minh City.

\end{document}